\documentclass{article}
\usepackage{spconf,amsmath,epsfig}
\usepackage{graphicx}
\usepackage{amsmath,amsfonts}
\usepackage{algorithmic}
\usepackage{algorithm}
\usepackage{array}

\let\OLDthebibliography\thebibliography
\renewcommand\thebibliography[1]{
  \OLDthebibliography{#1}
  \setlength{\parskip}{0pt}
  \setlength{\itemsep}{0pt plus 0.3ex}
}

\begin{document}\sloppy

\def\x{{\mathbf x}}
\def\L{{\cal L}}

\setlength{\abovedisplayskip}{3pt}
\setlength{\belowdisplayskip}{3pt}
\setlength{\abovedisplayshortskip}{3pt}
\setlength{\belowdisplayshortskip}{3pt}

\title{Cross-Attention is Not Always Needed: Dynamic Cross-Attention for Audio-Visual Dimensional Emotion Recognition}
%
\name{R. Gnana Praveen, Jahangir Alam}
\address{ Computer Research Institute of Montreal (CRIM), Canada}

\maketitle

\begin{abstract}
In video-based emotion recognition, audio and visual modalities are often expected to have a complementary relationship, which is widely explored using cross-attention. However, they may also exhibit weak complementary relationships, resulting in poor representations of audio-visual features, thus degrading the performance of the system. To address this issue, we propose Dynamic Cross-Attention (DCA) that can dynamically select cross-attended or unattended features on the fly based on their strong or weak complementary relationship with each other, respectively. Specifically, a simple yet efficient gating layer is designed to evaluate the contribution of the cross-attention mechanism and choose cross-attended features only when they exhibit a strong complementary relationship, otherwise unattended features. We evaluate the performance of the proposed approach on the challenging RECOLA and Aff-Wild2 datasets. We also compare the proposed approach with other variants of cross-attention and show that the 
proposed model consistently improves the performance on both datasets.
\end{abstract}
\section{Introduction}
\label{sec:intro}

Automatic recognition of human emotions is a challenging problem, 
\begin{figure}[!t]
\centering
\includegraphics[width=1.0\linewidth]{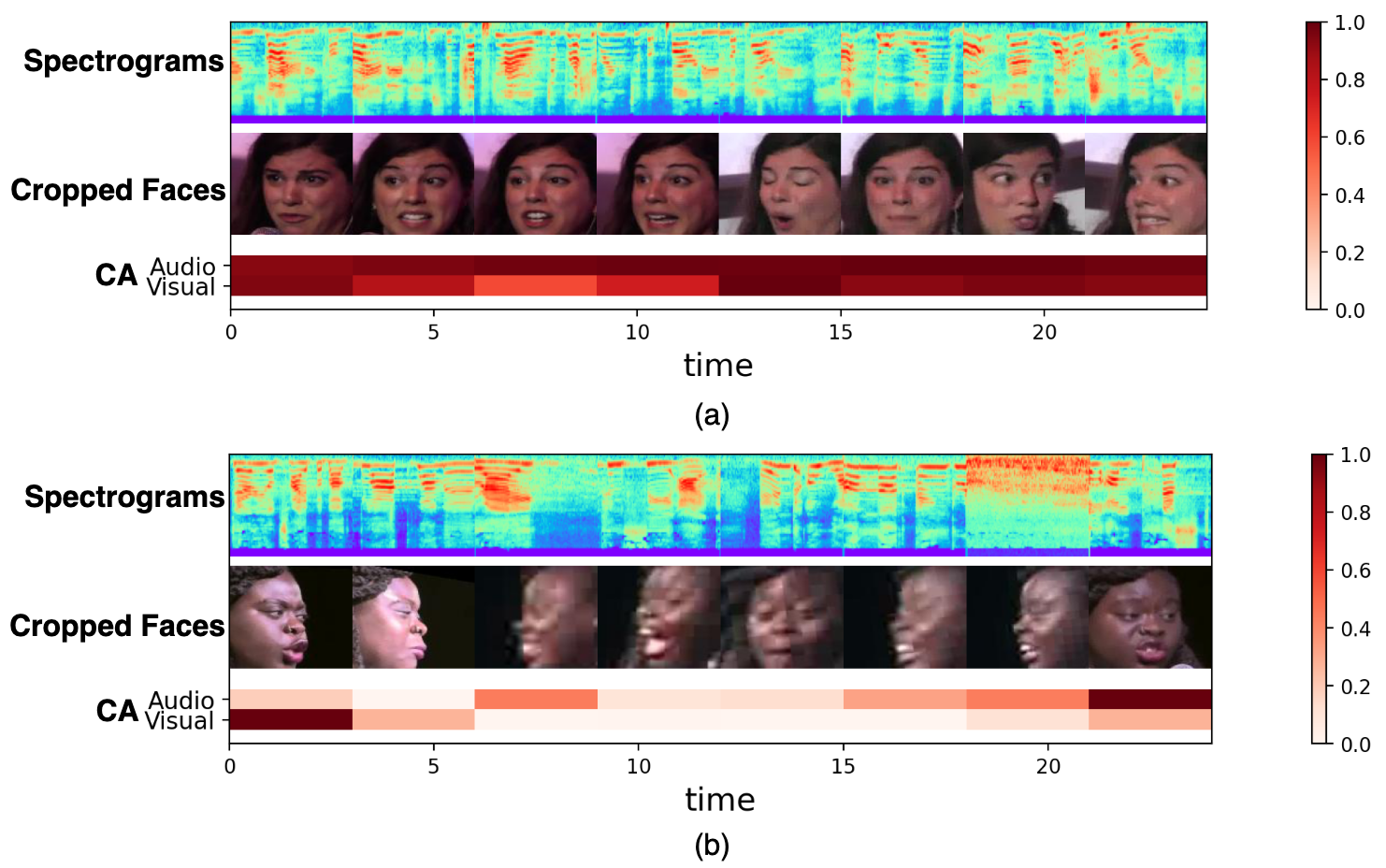}
\caption{\small (a) Attention scores based on cross-attention for the subject "12-24-1920x1080" from the validation set of the Aff-Wild2 dataset. Here, both audio and visual modalities strongly complement each other, thereby assigning higher attention scores for face and voice with significant expressions (b) Attention scores based on cross-attention for the subject "21-24-1920x1080" from the validation set of the Aff-Wild2 dataset. In this case, the facial modality is corrupted due to extreme blur and pose, however, it has rich vocal expressions. Attending the corrupted face to rich vocal expressions fails to assign higher attention scores for vocal expressions.}
\label{Demo of Complementary relationships}
\vspace{-3mm}
\end{figure}
typically formulated as the classification of emotions. 
However, humans can express a wide range of more subtle and complex emotions beyond them. 
In recent years, regression of expressions has gained a lot of attention because it has the potential to capture a wide range of expressions. 
Depending on the granularity of labels, the regression of emotions can be formulated as ordinal regression or continuous regression. Compared to ordinal regression, continuous (or dimensional) regression is even more challenging due to the highly complex process of obtaining annotations on continuous dimensions. Valence and arousal are widely used dimensions to estimate emotion intensities in a continuous domain. 
Valence reflects the wide range of emotions in the dimension of pleasantness, from being negative (sad) to positive (happy). In contrast, arousal spans a range of intensities, from passive (sleepiness) to active (high excitement). 
Recently, multimodal learning has achieved remarkable success by leveraging complementary relationships across the modalities using cross-modal attention for Emotion Recognition (ER) \cite{n20_interspeech,9667055}. The idea of cross-modal attention, often termed as Cross-Attention (CA) or co-attention, is to leverage one modality to attend to another modality based on cross-modal interactions \cite{10123038}.
However, audio and visual modalities may not always strongly complement each other, they may also exhibit weak complementary relationships \cite{9706879}.
It has been shown that the audio and visual modalities may also demonstrate conflicting (when one of the modalities is noisy or paradoxical) or dominating (when one of the modalities is restrained or unemotional) relationships for ER \cite{9552921}. 
When one of the modalities is noisy or restrained (weak complementary relationship), leveraging the noisy modality to attend to a good modality can deteriorate the fused Audio-Visual 
(A-V) feature representations \cite{9706879}. To better understand the problem of weak complementary relationships for ER, we provided an interpretability analysis by visualizing the attention scores (normalized between 0 and 1) of CA as shown in Fig. 
\ref{Demo of Complementary relationships}. We can observe that the audio and visual modalities demonstrate higher attention scores of CA for samples with strong complementary relationships (top image), where intense facial expressions are strongly complemented by intense vocal expressions. For weak complementary relationships, where intense vocal expressions are associated with noisy facial expressions (bottom image), we can see that both audio and visual modalities exhibit lower CA scores. Even though audio modality conveys strong emotional expressions, attending the audio with noisy visual lowers the attention scores for the audio modality. Therefore, CA fails to retain rich information of intense vocal expressions, resulting in poor A-V feature representations. Motivated by this insight, we have investigated the prospect of developing a robust model that can dynamically choose when to integrate the cross-attended or unattended features depending on the strong or weak complementary relationships, respectively.    

In this work, we propose a Dynamic Cross Attention (DCA) model that can dynamically adapt to both strong and weak complementary relationships by selecting the semantic features to deal with the problem of weak complementary relationships. Specifically, we introduce a gating layer to evaluate the strength of the complementary relationships and select the cross-attended or unattended features dynamically based on strong or weak complementary relationships, respectively.    
Therefore, the proposed DCA model adds more flexibility to the CA framework and improves the fusion performance even when the modalities exhibit weak complementary relationships. 
The proposed model is a simple, yet efficient way of dynamically selecting the most relevant features, which can also be adapted to any variant of the CA model. 
To our knowledge, this is the first work to investigate incompatibility issues (weak complementary relationships) between audio and visual modalities for ER. 

The major contributions of the paper can be summarized as follows. (i) We investigate the potential of the CA model in leveraging complementary relationships and show that weak complementary relationships degrade the fusion performance. 
(ii) We propose a DCA model to dynamically select the cross-attended or unattended features based on the strength of their complementary relationships across audio and visual modalities. (iii) The proposed model is further evaluated on different variants of CA and demonstrated that the proposed model consistently improves the performance of the system on both RECOLA and Aff-Wild2 datasets.


\section{Related Work}

\noindent \textbf{Attention models for ER:} 
Recently, multimodal transformers with CA 
showed significant improvement for ER \cite{Karas_2022_CVPR,Zhou_2023_CVPR,9857097}. 
Parthasarathy et al. \cite{srini_2021_SLT} explored multimodal transformers, where the CA module is integrated with the self-attention module to obtain the A-V cross-modal feature representations. 
Zhang et al. \cite{9607460} 
proposed a leader-follower attention mechanism by considering the visual modality as the primary channel, while the audio modality is used as a supplementary channel to boost visual performance. 
Karas et al. \cite{Karas_2022_CVPR} and Meng et al. \cite{9857097} showed improvement in fusion performance by exploring a set of fusion models based on LSTMs and transformers. 
Zhou et al. \cite{Zhou_2023_CVPR} 
explored temporal convolutional networks (TCNs) for individual modalities, whereas Zhang et al \cite{Zhang_2023_CVPR} exploited masked auto-encoders for visual modality.   
However, most of these methods \cite{Zhou_2023_CVPR,Zhang_2023_CVPR,Karas_2022_CVPR,9857097} 
rely on a naive fusion approach or ensemble-based fusion using transformers and LSTMs. 
Unlike these approaches, 
Praveen et al. \cite{9667055} proposed a CA model to effectively leverage complementary relationships by allowing the modalities to interact with each other. They have extended the approach by introducing joint feature representation in the CA framework to capture intermodal and intramodal relationships \cite{10005783} and recursive fusion \cite{praveen2023recursive}. 
Although these methods 
have shown impressive performance with CA, they rely on the assumption that audio and visual modalities always exhibit strong complementary relationships. When audio and visual modalities exhibit a weak complementary relationship due to restrained or noisy modalities, these methods will result in poor performance.

\begin{figure*}[!t]
\centering
\includegraphics[width=0.8\linewidth]{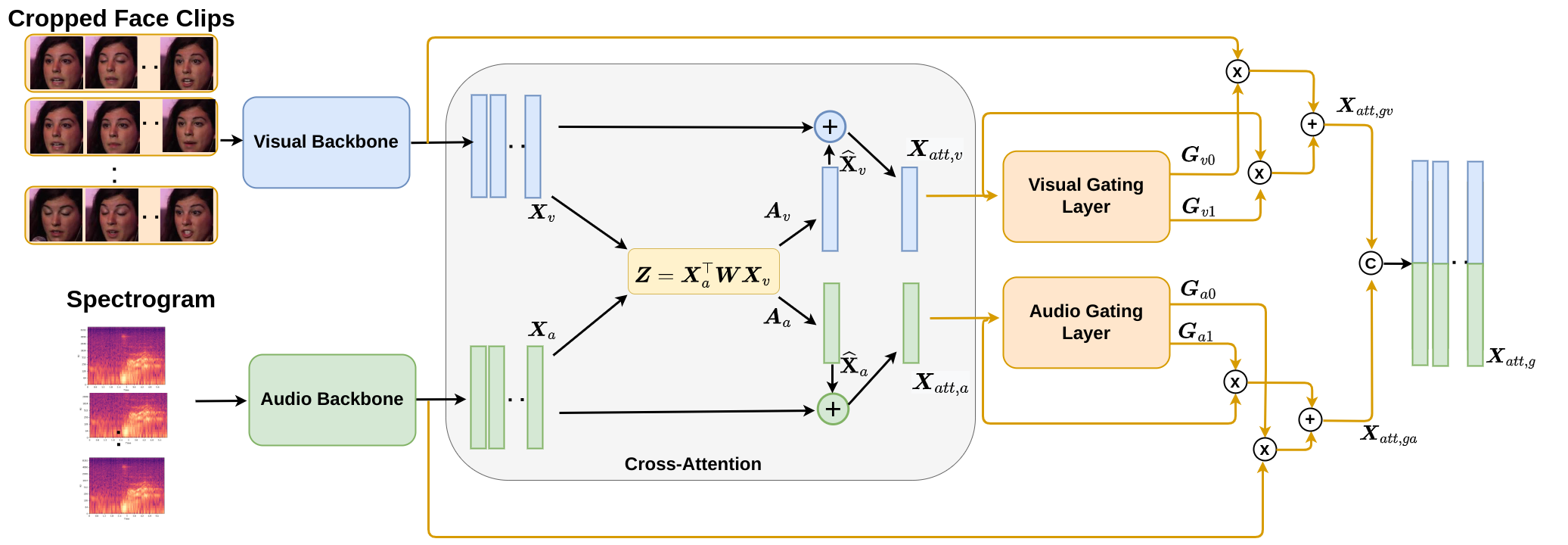}
\caption{\small Illustration of the proposed Dynamic Cross-Attention (DCA) model with vanilla Cross-Attention (CA) as the baseline.}
\label{AdapJA}
\vspace{-3mm}
\end{figure*}
\noindent \textbf{Gating-Based Attention:} Conventionally gating mechanisms have been explored for multimodal fusion to control the flow of modalities to reduce redundancy \cite{Xue_Li_Zhang_2020} or to mitigate the impact of noisy modalities \cite{9053012,10096438,DU2022108107}.
Aspandi et al. \cite{9726856} proposed a gated-sequence neural network for dimensional ER, where the gating mechanism is explored for temporal modeling 
to adaptively fuse the modalities based on their relative importance. 
Kumar et al. \cite{9053012} explored the conditional gating mechanism using a nonlinear transformation by modulating the cross-modal interactions to learn the relative importance of modalities. 
Liu et al. \cite{liu20b_interspeech} used a two-stage gating mechanism to control the alignment and contribution of the modalities for speech and text. 
Unlike these approaches, we have focused on handling the problem of weak complementary relationships using a conditional gating mechanism 
by dynamically selecting the most relevant features based on strong or weak complementary relationships, respectively.

\section{Proposed Model}
\noindent \textbf{A) Notations:}
For an input video sub-sequence $S$, $L$ non-overlapping video clips are uniformly sampled, and the corresponding deep feature vectors are obtained from the pre-trained models of audio and visual modalities. Let ${\boldsymbol X}_{a}$ and ${\boldsymbol X}_{v}$ denote the deep feature vectors of audio and visual modalities, respectively for the given input video sub-sequence $S$ of fixed size, which is expressed as 
${ \boldsymbol X}_{a}=  \{  x_{a}^1,  x_{a}^2, ...,  x_{ a}^L \} \in \mathbb{R}^{d_a\times L}$ 
and 
${ \boldsymbol X}_{v}=  \{  x_{ v}^1,  x_{ v}^2, ...,  x_{ v}^L \} \in \mathbb{R}^{d_v\times L}$
where ${d_a}$ and ${d_v}$ represent the dimensions of the audio and visual feature vectors, respectively, and $ x_{ a}^{ l}$ and $ x_{ v}^{ l}$ denotes the audio and visual feature vectors of the video clips, respectively, for $l = 1, 2, ..., L$ clips.

\noindent \textbf{B) Preliminary-Cross Attention:}
In this section, we briefly introduce the cross-attention (CA) \cite{9667055} (baseline fusion model) as a preliminary to the proposed model.  
Given the audio and visual feature vectors ${\boldsymbol X}_{ a}$ and ${\boldsymbol X}_{ v}$ for a video sub-sequence $S$, the cross-correlation across the modalities are computed as 
$\boldsymbol Z={\boldsymbol X}_{ a}^\top \boldsymbol W{\boldsymbol X}_{ v}$
where 
$\boldsymbol W\in\mathbb{R}^{d_a\times d_v}$ represents cross-correlation weights among the audio and visual features. 
Next, we compute the CA weights of audio and visual features, ${\boldsymbol A}_{ a}$ and ${\boldsymbol A}_{ v}$ by applying column-wise softmax of $\boldsymbol Z$ and $\boldsymbol Z^\top$, respectively. 
After obtaining the CA weights, they are used to obtain the attention maps of the audio and visual features as 
\begin{align}
\widehat{\mathbf{X}_{a}}=\mathbf{X}_{a}{{\mathbf A}_{ a}} \quad  \text{and} \quad
\widehat{\mathbf{X}_v}=\mathbf X_v{{\mathbf A}_{ v}}
\end{align}
The attention maps are added to the corresponding features to obtain the final cross-attended features ${\boldsymbol X}_{att,a}$ and ${\boldsymbol X}_{att,v}$.

\noindent \textbf{C) Dynamic Cross-Attention Model:}
The proposed DCA model is depicted in Fig. \ref{AdapJA} with vanilla CA as the baseline.
Given the cross-attended and unattended features of audio and visual modalities, we design a gating layer using a fully connected layer for each modality separately to obtain the attention weights for the attended and unattended features, which are given by 
\begin{align}
{\boldsymbol Y}_{go,v} = {\boldsymbol X}_{ att,v}^\top \boldsymbol W_{gl,v}
\hspace{-2mm} \quad \text{and} \hspace{-2mm} \quad {\boldsymbol Y}_{go,a} = {\boldsymbol X}_{att,a}^\top \boldsymbol W_{gl,a}
\end{align}
where $\boldsymbol W_{gl,a} \in\mathbb{R}^{d_a\times 2}$, $\boldsymbol W_{gl,v}\in\mathbb{R}^{d_v\times 2}$ are the learnable weights of the gating layers and $\boldsymbol Y_{go,a} \in\mathbb{R}^{L\times 2}$, $\boldsymbol Y_{go,v} \in\mathbb{R}^{L\times 2}$ are outputs of gating layers of audio and visual modalities, respectively. To obtain probabilistic attention scores, the output of the gating layers is activated using a softmax function with a small temperature $T$, as given by 
\begin{align}
{\boldsymbol G}_{{a}}
=\frac{ e^{{\boldsymbol Y}_{go,a}/T}}{\overset{ K}{\underset{ k\boldsymbol=1}{\sum}} e^{{\boldsymbol Y}_{go,a}/T}}    
\quad  \text{and} \quad
{\boldsymbol G}_{{v}}
=\frac{ e^{{\boldsymbol Y}_{go,v}/T}}{\overset{K}{\underset{ k\boldsymbol=1}{\sum}} e^{{\boldsymbol Y}_{go,v}/T}}   
\label{softmax}
\end{align}
where $\boldsymbol {G}_{a} \in\mathbb{R}^{L\times 2}, \boldsymbol {G}_{v}\in\mathbb{R}^{L\times 2}$ denotes the probabilistic attention scores of audio and visual modalities. $K$ denotes the number of output units of the gating layer, which is $2$, one for cross-attended features and one for unattended features.  
These gating weights allow the selection of attended or unattended features dynamically based on the strong or weak complementary relationships, respectively. 
The parameter $T$ acts as a softmax temperature on the output of the gating layer to provide a regularization effect. In our experiments, we have empirically set the value of $T$ to 0.1 (see supplementary material for ablation study on $T$). 
Ideally, the gating output for unattended features is $0$ for samples with strong complementary relationships (vice-versa), allowing only the cross-attended features, which is the same as the CA model \cite{9667055}. 
By using a small temperature $T$, the non-selected unattended feature, which acts as a noisy signal gets a small weightage, thereby providing a regularization effect 
\cite{10.5555/3295222.3295264}. 
The probabilistic attention scores of the gating layer help to estimate the relevance of attended and unattended features for the accurate prediction of valence and arousal. The two columns of $\boldsymbol G_a$ correspond to the probabilistic attention scores of unattended features (first column) and cross-attended features (second column). To multiply with the corresponding feature vectors, each column is replicated to match the dimension of the corresponding feature vectors, which is denoted by $\boldsymbol G_{a0}$, $\boldsymbol G_{a1}$ and $\boldsymbol G_{v0}$, $\boldsymbol G_{v1}$ for audio and visual modalities respectively. Now, the replicated attention scores  
are multiplied with the corresponding cross-attended and unattended features of the respective modalities, which is further fed to the ReLU activation function as: 
\begin{align}
{\boldsymbol X}_{att,gv} = \text{ReLU}(\boldsymbol X_{v} \otimes \boldsymbol G_{v0} + {\boldsymbol X}_{att,v} \otimes \boldsymbol G_{v1}) \\ 
{\boldsymbol X}_{att,ga} = \text{ReLU}(\boldsymbol X_{a} \otimes \boldsymbol G_{a0} + {\boldsymbol X}_{att,a} \otimes \boldsymbol G_{a1})  
\end{align}
where $\otimes$ denotes element-wise multiplication. ${\boldsymbol X}_{att,ga}$ and ${\boldsymbol X}_{att,gv}$ denote the final attended feature vectors of audio and visual modalities, respectively obtained from the DCA model, which is further 
concatenated to obtain the A-V feature representation ${\boldsymbol X}_{att,g}$, followed by regression layers to get final predictions. 


\section{Results and Discussion}
\noindent \textbf{A) Datasets:} The proposed approach has been evaluated on the RECOLA \cite{6553805} and Aff-Wild2 \cite{10208745} datasets. The RECOLA dataset consists of 9.5 hours of multimodal recordings, recorded by 46 French-speaking participants. 
Most existing approaches \cite{cite6,cite7} have been evaluated on the dataset used for the AVEC 2016 challenge, which consists of 9 subjects for training and 9 subjects for validation. Therefore, we also validated the proposed approach on the AVEC 2016 challenge dataset. The Aff-Wild2 corpus is the largest dataset in the field of affective computing, consisting of $594$ videos collected from YouTube. 
Sixteen of these videos display two subjects, both of which have been annotated. In total, there are $2,993,081$ frames with $584$ subjects. 
The dataset is divided into 360, 72, and 162 videos in training, validation, and test sets, respectively in a subject-independent manner. For both datasets, annotations for valence and arousal are provided continuously in the range of $\lbrack-1,1\rbrack$. 

\noindent \textbf{B) Implementation Details:}
The audio and visual modalities are preprocessed and the feature vectors are obtained by following similar procedure to that of the baselines \cite{9667055,10005783,praveen2023recursive}. Following the baselines, we have used R3D network for visual modality and 
Resnet-18 for audio modality (see supplementary material for more implementation details).
Concordance Correlation Coefficient (CCC) is used as an evaluation metric, which has been widely used in the literature of dimensional ER to measure the level of agreement between the predictions and ground truth of valence and arousal \cite{cite7,9857097,praveen2023recursive,10005783}.

\begin{table}
  \caption{\small Ablation study for the impact of the DCA model on different baselines of the validation set on Aff-Wild2 dataset.}
  \label{tab:GAonDifferentBL}
  \centering
  \begin{tabular}{|c|c|c|c|}
 \hline
    \textbf{Method}&\textbf{Valence}&\textbf{Arousal}\\
 \hline
    CA \cite{9667055} w/o DCA & 0.541 & 0.517\\
    \hline
    CA \cite{9667055} w/ DCA & 0.624 & 0.582\\
    \hline
    TCA \cite{srini_2021_SLT} w/o DCA & 0.564 & 0.543 \\
    \hline
    TCA \cite{srini_2021_SLT} w/ DCA & 0.635 & 0.621 \\
    \hline
    JCA \cite{10005783} w/o DCA  & 0.657 & 0.580 \\
    \hline
    JCA \cite{10005783} w/ DCA  & 0.679 & 0.612\\
    \hline
    RJCA \cite{praveen2023recursive} w/o DCA & 0.721 & 0.694 \\
    \hline
    RJCA \cite{praveen2023recursive} w/ DCA & 0.742 & 0.718\\
     \hline
\end{tabular}
\vspace{-3mm}
\end{table}

\noindent \textbf{C) Ablation Study:}
We have conducted a series of experiments on the validation set of the Aff-Wild2 dataset to analyze the impact of the proposed DCA model on different 
variants of CA model and the results are presented in Table \ref{tab:GAonDifferentBL}. More specifically, we have considered four baseline models: Cross-Attention (CA) \cite{9667055}, Joint Cross-Attention (JCA) \cite{10005783}, Recursive Joint Cross-Attention (RJCA) \cite{praveen2023recursive}, and Transformer-based Cross-Attention (TCA) \cite{srini_2021_SLT}. 
First, we have implemented a simple baseline of the CA model to capture the complementary relationship between audio and visual modalities. Then we added the proposed DCA model to analyze the impact on the CA model, which has significantly improved the performance of the system. Similarly, we have also analyzed the impact of the proposed DCA model on other baseline models JCA, RJCA, and TCA. 
In all of these baselines, the fused A-V feature representations are influenced by the cross-modal interactions by using one modality to attend to the other modality (assuming strong complementary relationships). 
Since the proposed DCA model addresses the problem of weak complementary relationships, we can observe that the proposed model consistently improves the performance over all baselines.
In particular, the proposed DCA model shows a significant improvement with CA and TCA over that of JCA and RJCA. This can be due to the fact that CA and TCA models rely only on cross-modal attention, where audio modality is 
used to attend to visual modality and vice-versa. 
In the case of JCA and RJCA, the attention weights for each modality are based on both intra- and inter-modal relationships using the joint feature representation, thereby reducing the impact of noisy or restrained modality. We have also observed a similar trend of performance improvement for the ablation study on the RECOLA dataset, which is provided in supplementary material.
\begin{table}[!t]
\small
\centering
  \caption{\small CCC Performance of the proposed model with comparison to state-of-the-art methods on the validation set of RECOLA. 
  }
  \label{tab:comparisiontoSOAonRECOLA}
  \begin{tabular}{|c|c|c|c|}
  \hline
    \textbf{Method} & \textbf{Type of Fusion}  &\textbf{Valence}&  \textbf{Arousal} \\
\hline
	Tzirakis et al. \cite{cite7} & LSTM &  0.502 & 0.731 \\ 
 \hline
	Ortega et al. \cite{8914655}  &  Feature Concat &0.565 &  0.749  \\
 \hline
    Schoneveld et al. \cite{cite6}  & LSTM & 0.630 & 0.810  \\
    \hline
	 Praveen et al \cite{9667055} & CA &  0.687 & 0.831\\ 
  \hline
	Praveen et al.\cite{10005783}  & JCA & 0.728  & 0.842\\ 
 \hline
 Praveen et al.\cite{praveen2023recursive}  & RJCA &  0.736 & 0.855 \\ 
 \hline
    Ours & RJCA + DCA  & \textbf{0.743}  & \textbf{0.867} \\
\hline
  \end{tabular}
  \vspace{-3mm}
\end{table}
\begin{table*}[!h]
\centering
  \caption{\small CCC Performance of the proposed model in comparison to state-of-the-art methods on the Aff-Wild2. 
  }
  \label{tab:comparisiontoSOAonAffwild2}
  \begin{tabular}{|c|c|c|c|c|c|c|c|}
 \hline
 \textbf{Method }& 
 \textbf{Type of } & \multicolumn{2}{|c|}{\textbf{Validation Set}} & \multicolumn{2}{|c|}{\textbf{Test Set}} \\

    \cline{3-6}
     & \textbf{Fusion} & \textbf{Valence} & \textbf{Arousal} & \textbf{Valence} & \textbf{Arousal} \\
      \hline
    Zhang et al. \cite{9607460} & Leader Follower Attention & 0.469 &  0.649 & 0.463 & 0.492  \\
          \hline
    Karas et al. \cite{Karas_2022_CVPR} & LSTM + Transformers & 0.388 & 0.551 & 0.418 & 0.407 \\
          \hline
    Meng et al. \cite{9857097} & LSTM + Transformers & 0.605 & 0.668 & \textbf{0.606} & \textbf{0.596} \\
          \hline
     Zhou et al. \cite{Zhou_2023_CVPR} & TCN + Transformers &0.550 & 0.680 & 0.566 & 0.500\\
     \hline
    Zhang et al \cite{Zhang_2023_CVPR} &  Transformers &0.648 & 0.705 & 0.523 & 0.545\\
    \hline
    Praveen et al.\cite{10005783}& JCA & 0.657  & 0.580 & 0.451 & 0.389\\
    \hline
    Praveen et al.\cite{praveen2023recursive} &
    RJCA & 0.721 & 0.694 & 0.467 & 0.405\\
    \hline
    Ours & RJCA + DCA & \textbf{0.742} &  \textbf{0.718} & \textbf{0.507} & \textbf{0.473}\\
    \hline
  \end{tabular}
  \vspace{-3mm}
\end{table*}     

\noindent \textbf{E) Comparison to state-of-the-art:}
In all these experiments, we performed multiple runs and took an average of three runs to have statistically stable results. 

\noindent \textbf{RECOLA: }Table \ref{tab:comparisiontoSOAonRECOLA} shows the performance comparison of the proposed approach with the relevant approaches on the development set of the RECOLA dataset. Since the test set is no longer supported for RECOLA, we have evaluated only the development set. Although Tzirakis et al. \cite{cite7} explored deep features with LSTM-based fusion 
on the concatenated features, the fusion performance is not better than that of individual modalities. 
The performance has been improved by Ortega et al. \cite{8914655} using pre-trained models on FER2013 dataset for visual, and Low-Level Descriptors (LLD) for audio. 
Schoneveld et al. \cite{cite6} further improved the fusion performance using knowledge distillation for visual features, and a VGG network for audio features. 
However, these methods fail to effectively capture the complementary inter-modal relationships. Unlike these approaches, Praveen et al. \cite{9667055,10005783,praveen2023recursive} explored CA to capture the complementary relationship across audio and visual modalities and showed improvement over prior methods. 
Although \cite{9667055,10005783,praveen2023recursive} achieved superior performance, they failed to deal with weak complementary relationships. By deploying the proposed DCA model, we have achieved a further improvement in the fusion performance for both valence and arousal.
\begin{figure}[!t]
\hspace{-15mm}
\includegraphics[width=1.2\linewidth]{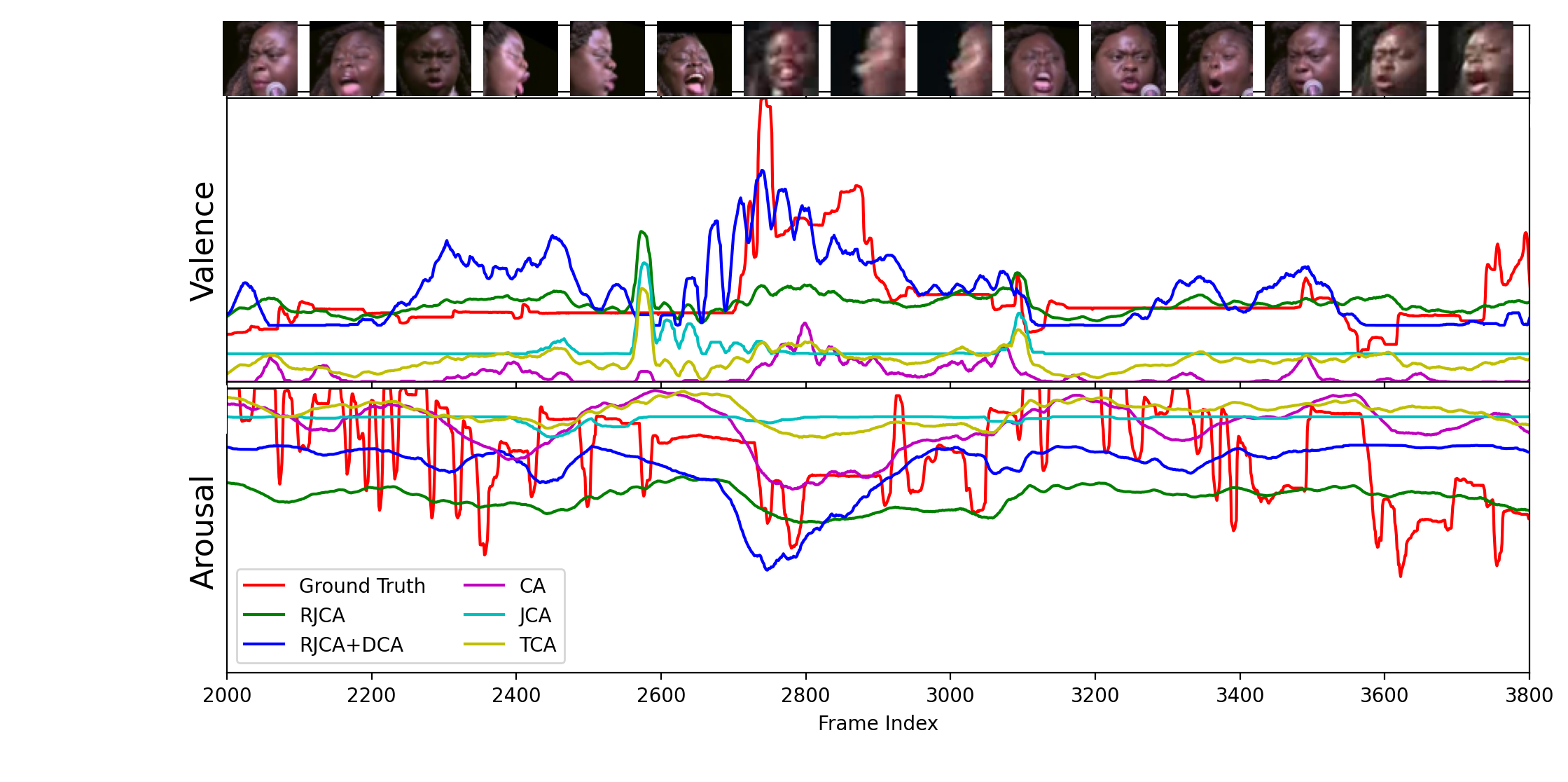}
\vspace{-10mm}
\caption{\small Visualization of the predictions of valence and arousal for subjects "21-24-1920x1080" of the validation set of Aff-Wild2 dataset.}
\label{fig:visualization}
\vspace{-3mm}
\end{figure}

\noindent \textbf{Aff-Wild2: }Table \ref{tab:comparisiontoSOAonAffwild2} shows the performance of the proposed approach against the relevant state-of-the-art A-V fusion models on both development and test set partitions of the Aff-Wild2 dataset. Most of the related work on this dataset has been submitted to the 
ABAW challenges \cite{10208745}. Therefore we compare the proposed approach with the relevant state-of-the-art models of ABAW challenges for A-V fusion in dimensional ER. 
Zhang et al. \cite{9607460} exploited audio as a supplementary channel to boost the performance of visual modality.  
Meng et al. \cite{9857097} 
showed significant improvement in both development and test sets by leveraging three external datasets and multiple backbones for audio and visual modalities using an ensemble of LSTMs and transformers. Similarly, Zhou et al. \cite{Zhou_2023_CVPR} and Zhang et al. \cite{Zhang_2023_CVPR} also explored multiple backbones to achieve better generalization and improved performance on the test set. However, both of these methods \cite{Zhou_2023_CVPR} and \cite{Zhang_2023_CVPR} rely on a naive fusion approach based on transformers. Most approaches \cite{Zhang_2023_CVPR,Zhou_2023_CVPR,9857097} explored multiple sophisticated backbones with ensemble models and external datasets to attain better generalization ability and improve test set performance. 

Unlike these approaches, Praveen et al. \cite{10005783} focused on improving the fusion performance by effectively capturing both intra- and inter-modal relationships across the audio and visual modalities by deploying a joint representation in the CA framework. 
They further improved the fusion performance by introducing LSTMs and recursive fusion to obtain robust A-V feature representations \cite{praveen2023recursive}. Though \cite{10005783} and \cite{praveen2023recursive} do not outperform the performance on the test set compared to that of \cite{Zhou_2023_CVPR,Zhang_2023_CVPR,9857097}, their performance can be solely attributed by the sophisticated fusion models based on CA as they use only single backbones for each modality without any external datasets. Since the primary focus of our work is to improve the fusion performance by handling weak complementary relationships, we have evaluated the proposed model on these sophisticated CA-based fusion models \cite{praveen2023recursive}. By deploying the proposed DCA model on the RJCA model \cite{praveen2023recursive}, we can observe that the fusion performance has been further improved 
on both development and test sets. The performance improvement is more emphasized in the test set, which can be attributed to the regularization effect provided by the variable $T$ in Eq. \ref{softmax}. To have a fair comparison, we also compared the fusion models (ensemble of transformers and LSTMs) of other state-of-the-art methods \cite{Zhou_2023_CVPR,Zhang_2023_CVPR,9857097} using our backbones (see supplementary material). Since the Aff-Wild2 dataset is obtained from extremely challenging environments, the videos are highly vulnerable to exhibiting weak complementary relationships. Therefore, the improvement in fusion performance of the proposed DCA model is relatively better than that of the RECOLA dataset.

\noindent \textbf{F) Qualitative Evaluation:}
We also provided interpretability analysis for a better understanding of the proposed model using the visualization of the predictions of valence and arousal of proposed model along with the considered baselines (as shown in Fig. \ref{fig:visualization}). We have chosen the subject named "21-24-1920x1080" from the validation set of the Aff-Wild2 dataset to demonstrate the superiority of the proposed model in handling weak complementary relationships. 
All the baseline models can track the ground truth of the dimensional emotions by leveraging the complementary relationships across the modalities. Since JCA and RJCA exploit both intra- and inter-modal relationships using joint representation, they are able to perform relatively better than CA and TCA by closely tracking the ground truth.  
However, all these baseline models do not deal with the problem of weak complementary relationships between the audio and visual modalities. Since the proposed DCA model handles the problem of weak complementary relationships while retaining the potential of strong complementary relationships, the proposed DCA model is able to effectively track the ground truth better than all the baseline models. As we can observe in Fig. \ref{fig:visualization}, even though the facial expressions of some of the frames (2600 to 3000) are corrupted due to extreme pose and blur, the proposed DCA model is still able to closely track the variations of the ground truth by leveraging the rich vocal expressions without being contaminated. All the baseline models fail to track the variations in the ground truth as the rich vocal expressions are contaminated by the noisy visual modality, especially for valence (see more visualizations in supplementary material). Since we perform smoothening operations on predictions, they do not follow sudden fluctuations in ground truth.

\section{Conclusion}
In this paper, we investigated the issues with weak complementary relationships across audio and visual modalities in the framework of CA for ER. 
To address this issue, we proposed a simple, yet efficient DCA model to effectively capture the inter-modal relationships by handling the problem of weak complementary relationships while retaining the benefit of strong complementary relationships.
By adaptively choosing the most relevant features of the individual modalities based on the gated attention scores, the proposed model is able to adapt to both strong and weak complementary relationships. Experimental results demonstrate the superiority of the proposed model on the considered baseline models on two datasets. By levering advanced backbones for audio and visual modalities, we can also expect significant improvement in fusion performance on test sets.

\small
\bibliographystyle{IEEEbib}
\bibliography{icme2023template}

\begin{thebibliography}{10}

\bibitem{n20_interspeech}
Krishna~D. N. and Ankita Patil,
\newblock ``Multimodal emotion recognition using cross-modal attention and 1d convolutional neural networks,''
\newblock in {\em Interspeech}, 2020.

\bibitem{9667055}
R.~Gnana Praveen, Eric Granger, and Patrick Cardinal,
\newblock ``Cross attentional audio-visual fusion for dimensional emotion recognition,''
\newblock in {\em FG}, 2021.

\bibitem{10123038}
Peng Xu, Xiatian Zhu, and David~A. Clifton,
\newblock ``Multimodal learning with transformers: A survey,''
\newblock {\em TPAMI}, 2023.

\bibitem{9706879}
Jun-Tae Lee, Sungrack Yun, and Mihir Jain,
\newblock ``Leaky gated cross-attention for weakly supervised multi-modal temporal action localization,''
\newblock in {\em WACV}, 2022.

\bibitem{9552921}
Xingbo Wang, Jianben He, Zhihua Jin, Muqiao Yang, Yong Wang, and Huamin Qu,
\newblock ``M2lens: Visualizing and explaining multimodal models for sentiment analysis,''
\newblock {\em TVCG}, 2022.

\bibitem{Karas_2022_CVPR}
Vincent Karas, Mani~Kumar Tellamekala, Adria Mallol-Ragolta, Michel Valstar, and Bj\"orn~W. Schuller,
\newblock ``Time-continuous audiovisual fusion with recurrence vs attention for in-the-wild affect recognition,''
\newblock in {\em CVPRW}, 2022.

\bibitem{Zhou_2023_CVPR}
Weiwei Zhou, Jiada Lu, Zhaolong Xiong, and Weifeng Wang,
\newblock ``Leveraging tcn and transformer for effective visual-audio fusion in continuous emotion recognition,''
\newblock in {\em CVPRW}, 2023.

\bibitem{9857097}
Liyu Meng, Yuchen Liu, Xiaolong Liu, Zhaopei Huang, Wenqiang Jiang, Tenggan Zhang, Chuanhe Liu, and Qin Jin,
\newblock ``Valence and arousal estimation based on multimodal temporal-aware features for videos in the wild,''
\newblock in {\em CVPRW}, 2022.

\bibitem{srini_2021_SLT}
Srinivas Parthasarathy and Shiva Sundaram,
\newblock ``Detecting expressions with multimodal transformers,''
\newblock in {\em IEEE SLT Workshop}, 2021.

\bibitem{9607460}
Su~Zhang, Yi~Ding, Ziquan Wei, and Cuntai Guan,
\newblock ``Continuous emotion recognition with audio-visual leader-follower attentive fusion,''
\newblock in {\em ICCVW}, 2021.

\bibitem{Zhang_2023_CVPR}
Wei Zhang, Bowen Ma, Feng Qiu, and Yu~Ding,
\newblock ``Multi-modal facial affective analysis based on masked autoencoder,''
\newblock in {\em IEEE CVPRW}, 2023.

\bibitem{10005783}
R~Gnana Praveen, Patrick Cardinal, and Eric Granger,
\newblock ``Audio-visual fusion for emotion recognition in the valence-arousal space using joint cross-attention,''
\newblock {\em TBIOM}, 2023.

\bibitem{praveen2023recursive}
R~Gnana Praveen, Eric Granger, and Patrick Cardinal,
\newblock ``Recursive joint attention for audio-visual fusion in regression based emotion recognition,''
\newblock in {\em ICASSP}, 2023.

\bibitem{Xue_Li_Zhang_2020}
Lanqing Xue, Xiaopeng Li, and Nevin~L. Zhang,
\newblock ``Not all attention is needed: Gated attention network for sequence data,''
\newblock {\em AAAI}, 2020.

\bibitem{9053012}
Ayush Kumar and Jithendra Vepa,
\newblock ``Gated mechanism for attention based multi modal sentiment analysis,''
\newblock in {\em ICASSP}, 2020.

\bibitem{10096438}
Weikuo Guo and Xiangwei Kong,
\newblock ``Embrace smaller attention: Efficient cross-modal matching with dual gated attention fusion,''
\newblock in {\em Proc. of IEEE ICASSP}, 2023.

\bibitem{DU2022108107}
Yongping Du, Yang Liu, Zhi Peng, and Xingnan Jin,
\newblock ``Gated attention fusion network for multimodal sentiment classification,''
\newblock {\em Knowledge-Based Systems}, 2022.

\bibitem{9726856}
Decky Aspandi, Federico Sukno, Bjorn~W. Schuller, and Xavier Binefa,
\newblock ``Audio-visual gated-sequenced neural networks for affect recognition,''
\newblock {\em TAC}, 2022.

\bibitem{liu20b_interspeech}
Pengfei Liu, Kun Li, and Helen Meng,
\newblock ``Group gated fusion on attention-based bidirectional alignment for multimodal emotion recognition,''
\newblock in {\em Interspeech}, 2020.

\bibitem{10.5555/3295222.3295264}
Hyeonwoo Noh, Tackgeun You, Jonghwan Mun, and Bohyung Han,
\newblock ``Regularizing deep neural networks by noise: Its interpretation and optimization,''
\newblock in {\em NIPS}, 2017.

\bibitem{6553805}
Fabien Ringeval, Andreas Sonderegger, Juergen Sauer, and Denis Lalanne,
\newblock ``Introducing the recola multimodal corpus of remote collaborative and affective interactions,''
\newblock in {\em FG}, 2013.

\bibitem{10208745}
Dimitrios Kollias, Panagiotis Tzirakis, Alice Baird, Alan Cowen, and Stefanos Zafeiriou,
\newblock ``Abaw: Valence-arousal estimation, expression recognition, action unit detection \& emotional reaction intensity estimation challenges,''
\newblock in {\em CVPRW}, 2023.

\bibitem{cite6}
Schoneveld Liam, Othmani Alice, and Abdelkawy Hazem,
\newblock ``Leveraging recent advances in deep learning for audio-visual emotion recognition,''
\newblock {\em PR Letters}, 2021.

\bibitem{cite7}
Panagiotis Tzirakis, George Trigeorgis, Mihalis~A. Nicolaou, Björn~W. Schuller, and Stefanos Zafeiriou,
\newblock ``End-to-end multimodal emotion recognition using deep neural networks,''
\newblock {\em JSTSP}, 2017.

\bibitem{8914655}
Juan D.~S. Ortega, Patrick Cardinal, and Alessandro~L. Koerich,
\newblock ``Emotion recognition using fusion of audio and video features,''
\newblock in {\em SMC}, 2019.

\end{thebibliography}

\end{document}